\documentclass[conference]{IEEEtran}
\IEEEoverridecommandlockouts
\usepackage{cite}
\usepackage{amsmath,amssymb,amsfonts}
\usepackage{graphicx}
\usepackage{booktabs}
\usepackage{multirow}
\usepackage{textcomp}
\usepackage{xcolor}
\usepackage{url}

\begin{document}
\bstctlcite{IEEEexample:BSTcontrol}

\title{MATS: A Multi-Agent Framework for\\Controllable and Reflective Tumor Synthesis%
\thanks{This work was supported in part by the National Natural Science Foundation of China under Grant 82471933, in part by the Clinical Research Special Project of the Shanghai Municipal Health Commission under Grant 202340123, and in part by the China Postdoctoral Science Foundation under Grant 2025M771574.}}

\author{\IEEEauthorblockN{\parbox{0.96\textwidth}{\centering
Zhongnan Shi\textsuperscript{1},
Keyang Zhou\textsuperscript{1},
Siyi Li\textsuperscript{1},
Qichao Ying\textsuperscript{1},
Xinchen Ma\textsuperscript{2},\\
Shuyi Gong\textsuperscript{2},
Wei Tang\textsuperscript{2},
Sheng Li\textsuperscript{1},
Zhenxing Qian\textsuperscript{1, *}}}
\IEEEauthorblockA{\textsuperscript{1}Fudan University, Shanghai, China\\
\textsuperscript{2}Fudan University Shanghai Cancer Center, Shanghai, China\\
*Corresponding author}
}

\maketitle

\begin{abstract}
Synthesizing realistic tumors on medical images helps address data scarcity for clinical AI. Existing diffusion-based methods, however, lack fine-grained attribute-level controllability, closed-loop quality verification, and awareness of which morphological patterns are underrepresented in the target distribution. We present MATS, a multi-agent framework for text-controllable, cross-organ tumor synthesis. Given a target disease, MATS first profiles its morphological distribution over a fine-grained attribute space covering tumor geometry and appearance, and constructs a generation plan that amplifies underrepresented subtypes while keeping common ones dominant. It then localizes the target organ, proposes anatomically valid placement, and executes two-stage diffusion to produce the tumor mask and render texture, each conditioned on the corresponding attributes. A closed-loop reflective mechanism verifies attribute alignment at three checkpoints throughout synthesis and provides targeted feedback to guide re-generation when misalignment is detected. Trained exclusively on a single-source pancreatic neuroendocrine tumor dataset, MATS consistently improves downstream segmentation both in-domain and across LiTS, KiTS, and MSD-Pancreas benchmarks, whereas most prior generative augmentation methods fall below real-only training.
\end{abstract}


\section{Introduction}

Medical image datasets for tumor analysis~\cite{sonbol2022survival,chang2022progress} are often limited in scale and diversity, which constrains the performance of downstream clinical AI models. Tumor image synthesis has emerged as a practical strategy to enrich these datasets by generating realistic tumor samples on healthy scans. Effective synthesis, however, demands structured pathological attributes as the control axis~\cite{controlnet}, a verification mechanism that ensures generated tumors faithfully reflect the specified attributes, and awareness of which morphological patterns are missing in the target distribution.

Recent diffusion-based methods have advanced tumor synthesis considerably. DiffTumor~\cite{difftumor2024} achieves cross-organ generalization by leveraging shared imaging characteristics of early-stage tumors across organs. However, it does not support fine-grained morphological control, as generated shape and texture cannot be specified at the attribute level. LesionDiffusion~\cite{lesiondiffusion2025} introduces text-guided conditioning via structured lesion reports across multiple organs, yet its control space is coupled with organ-specific report formats and requires multi-organ training data with paired reports. Beyond controllability, existing synthesis pipelines generally lack closed-loop quality review. Multi-agent systems have shown promise in medical AI reasoning~\cite{pathfinder2025,mmedagent}, yet remain unexplored for controllable image generation. Moreover, few methods consider whether generated data addresses the specific morphological patterns that are underrepresented in the target dataset.

We propose a Multi-Agent Tumor Synthesis framework, or \textbf{MATS} in short, for controllable, reflective, and cross-organ tumor synthesis. The \textbf{Profiling Agent} profiles the target disease over a fine-grained attribute space covering tumor geometry and appearance, and constructs a generation plan that amplifies underrepresented subtypes while keeping common ones dominant, rather than naively inverting the long tail. The \textbf{Navigation Agent} localizes the target organ and proposes anatomically valid bounding boxes. The \textbf{Synthesis Agent} executes two-stage diffusion to generate the tumor mask and render texture, each conditioned on the corresponding attributes, with a cross-organ adapter for multi-organ transfer. The \textbf{Review Agent} checks spatial placement, mask geometry, and texture appearance at three checkpoints. When misalignment is detected, it sends targeted feedback to the responsible stage.

Because fine-grained morphological annotations are rare in public datasets, we contribute PNETS, 740 CT scans with expert masks, structured reports, and morphological labels, spanning solid, cystic, and mixed presentations with varying vascular patterns. The synthesis models are prepared on PNETS alone. Transfer to liver, kidney, and pancreas trains only a lightweight organ-appearance adapter. Experiments on PNETS, LiTS~\cite{lits2023}, KiTS~\cite{kits2023}, and MSD-Pancreas~\cite{msd2022} show gains over real-only segmentation on both backbones. MATS is best in 8 of 9 in-domain attribute groups; the isovascular group does not improve.

\section{Related Work}

\subsection{Controllable Tumor and Lesion Synthesis}
Generative models address data scarcity in medical imaging by synthesizing additional samples. Diffusion models~\cite{ho2020ddpm,rombach2022ldm} now generate realistic and controllable images. DiffTumor~\cite{difftumor2024} synthesizes tumors via latent diffusion over VQ-GAN~\cite{esser2021vqgan}, while FreeTumor~\cite{freetumor2025}, MedSegFactory~\cite{medsegfactory2025}, and MRGen~\cite{mrgen2025} scale synthesis or generate image-mask pairs. LeFusion~\cite{lefusion2025} decouples mask and appearance with histogram texture control but lacks semantic conditioning. LesionDiffusion~\cite{lesiondiffusion2025} conditions separate mask and texture networks on structured reports, yet treats generation as a single pass and relies on multi-organ paired supervision. ControlNet~\cite{controlnet} and IP-Adapter~\cite{ipadapter2023} inject structural signals into intermediate layers, but have not been combined with inference-time quality control for lesion synthesis. MATS adds that reflective check and a distribution-aware planner over a nine-dimensional attribute space.

\subsection{Multi-Agent Systems in Medical AI}
Multi-agent systems coordinate specialized modules through tool invocation and inter-module communication. In medical AI, PathFinder~\cite{pathfinder2025} orchestrates multiple agents for histopathology diagnosis, and MMedAgent~\cite{mmedagent} coordinates clinical tools. This paradigm fits staged synthesis workflows, yet remains underexplored in medical image generation. MATS assigns explicit responsibilities for distribution-aware profiling, anatomical navigation, two-stage diffusion synthesis, and closed-loop quality review.

\section{Method}\label{sec:method}

\begin{figure*}[t]
\centering
\includegraphics[width=\textwidth]{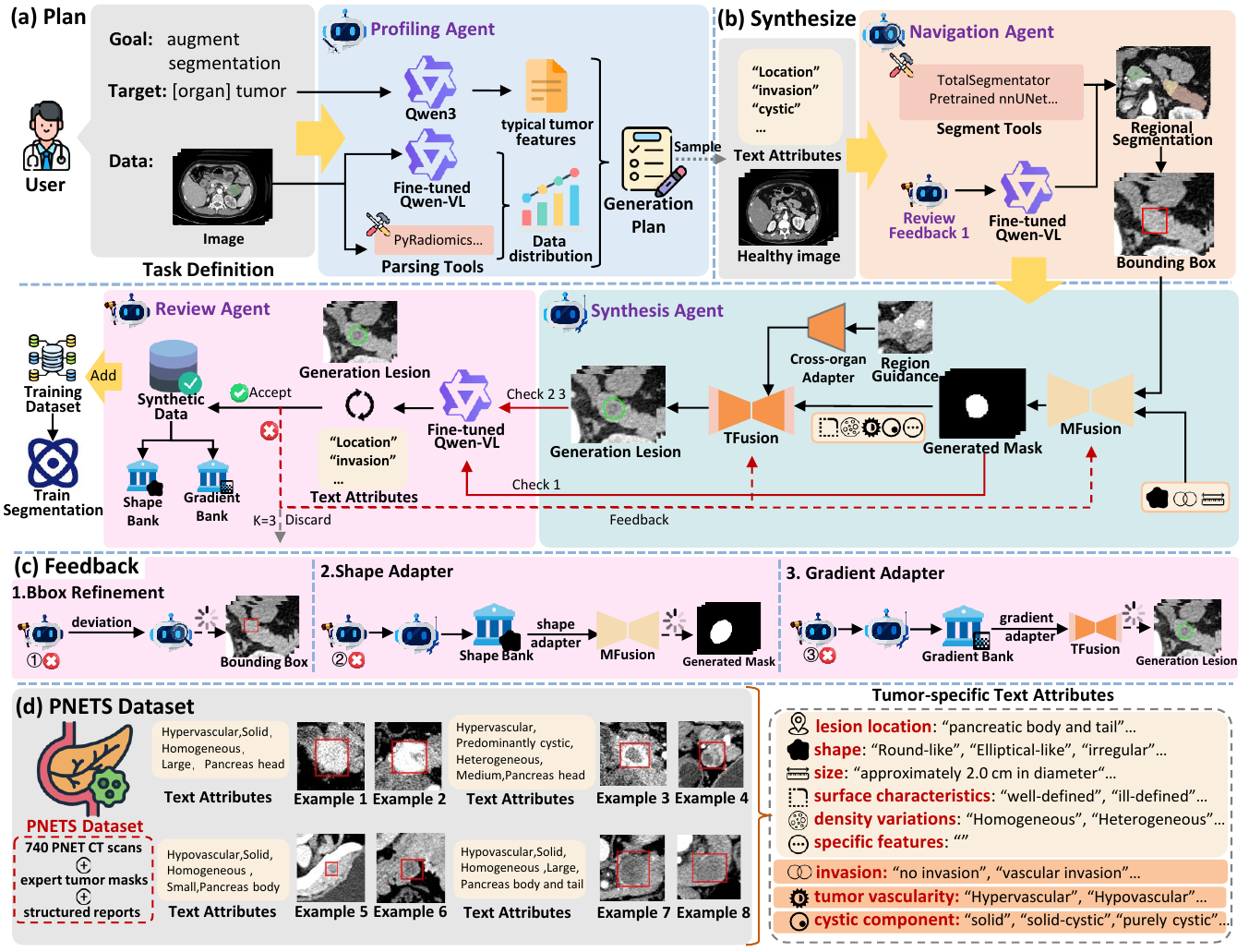}
\caption{MATS overview. \textbf{(a)}~Distribution-aware attribute profiling and generation planning. \textbf{(b)}~Sequential mask and image synthesis. \textbf{(c)}~Closed-loop feedback with corrective adapters. \textbf{(d)}~PNETS samples and morphological attributes.}
\label{fig:framework}
\end{figure*}

We formulate controllable tumor synthesis as a multi-agent coordination problem. Given a healthy scan $x$ and a target disease, MATS synthesizes a tumor image $\tilde{x}$ with mask $\tilde{m}$ that fills gaps in the disease's morphological attribute space. As shown in Fig.~\ref{fig:framework}, the Profiling Agent ($\mathcal{P}$) builds the generation plan, the Navigation Agent ($\mathcal{N}$) proposes the box, the Synthesis Agent ($\mathcal{S}$) runs MFusion and TFusion, and the Review Agent ($\mathcal{R}$) checks the three stages and can roll a failure back. Only $\mathcal{S}$ and its adapters contain trainable parameters. $\mathcal{P}$, $\mathcal{N}$, and $\mathcal{R}$ call frozen LLMs, a VLM, segmentors, and scripted checkers.

\subsection{PNETS Dataset for Agent Preparation}
MATS prepares all agents on one disease and transfers across organs, because cystic versus solid composition, vascularity, boundary regularity, and density heterogeneity are largely organ-agnostic. Morphological knowledge from one well-annotated source can therefore transfer without organ-specific tumor labels, provided the source covers this shared spectrum. Pancreatic neuroendocrine tumors do: they range from hypervascular to hypovascular and from solid to purely cystic.

PNETS contains 740 contrast-enhanced CT scans of histologically confirmed pancreatic neuroendocrine tumors~\cite{chang2022progress,sonbol2022survival}. Each case has an expert tumor mask, a structured radiology report, and categorical labels for the attributes defined below. Fine-grained PNET datasets remain scarce~\cite{icassp}. The synthesis split is 425 training, 174 validation, and 141 test cases. PNETS will be released under controlled access. All four agents are prepared on this split.

\subsection{Profiling Agent}
Each sample is $s_i=(I_i,r_i,m_i)$: a CT volume, an optional report, and a tumor mask. $\mathcal{P}$ profiles these cases over joint attributes used in clinical staging~\cite{lits2023} and sets the generation rate of each combination. The attribute is $\mathbf{A}=(\mathbf{A}_{\text{geo}},\mathbf{A}_{\text{app}})$, with geometry $\mathbf{A}_{\text{geo}}=(a_{\text{loc}},a_{\text{shape}},a_{\text{inv}},a_{\text{size}},a_{\text{surf}})$ for location, shape, invasion, size, and surface, and appearance $\mathbf{A}_{\text{app}}=(a_{\text{vas}},a_{\text{dens}},a_{\text{cyst}},a_{\text{spec}})$ for vascularity, density variation, cystic component, and specific features. Each coordinate is categorical. For example, $a_{\text{cyst}}$ is one of solid, solid-cystic, predominantly cystic, or purely cystic. A few categories per axis keep the space compact and clinically readable. Three tools read $\mathbf{A}$: (i)~Qwen3-8B parses the report, (ii)~a fine-tuned Qwen-VL~\cite{qwenvl} is used when the report is missing or incomplete, and (iii)~PyRadiomics measures geometry from the mask.

\noindent\textbf{Distribution-aware planning.} Let $p_{\text{real}}(\mathbf{A})$ be the empirical distribution of the target disease. Replicating it keeps the imbalance, and fully inverting it lets rare tuples overwhelm common ones. Following long-tailed learning~\cite{zhang2023longtail}, $\mathcal{P}$ targets
\begin{equation}
p_{\text{tgt}}(\mathbf{A})=\frac{p_{\text{real}}(\mathbf{A})^{\tau}}{\sum_{\mathbf{A}'}p_{\text{real}}(\mathbf{A}')^{\tau}},\quad\tau\in(0,1],
\label{eq:ptgt}
\end{equation}
which interpolates between the empirical distribution ($\tau{=}1$) and the uniform distribution ($\tau{\to}0$). For $N_{\text{real}}$ real training cases and a budget of $K$ synthetic cases, the allocation is
\begin{equation}
n_{\text{syn}}(\mathbf{A})=\max\!\big(0,\;(N_{\text{real}}+K)\,p_{\text{tgt}}(\mathbf{A})-N_{\text{real}}\,p_{\text{real}}(\mathbf{A})\big).
\label{eq:nsyn}
\end{equation}
If the positive counts sum to $S$, they are multiplied by $K/S$. Tuples already above the target receive no synthetic samples. At $\tau{=}1$, \eqref{eq:nsyn} reduces to $n_{\text{syn}}(\mathbf{A})=K\,p_{\text{real}}(\mathbf{A})$ and is well defined. We set $\tau{=}0.5$ and $K{=}N_{\text{real}}$. On a two-group $90/10$ distribution, \eqref{eq:ptgt} is $75/25$, \eqref{eq:nsyn} assigns $60$ and $40$ synthetic cases, and the combined set is $75/25$, so the original head remains the majority. Requests are drawn from this rescaled allocation. They are not drawn from $p_{\text{tgt}}$ itself.

\noindent\textbf{Conflict-aware filtering.} Some tuples are implausible, such as purely cystic, hypervascular, and severe vascular invasion together. A pairwise compatibility table, produced once by a domain LLM and checked by radiologists, sets those tuples to zero. The remaining counts are rescaled to $K$, and each request is sampled from the resulting categorical distribution.

\subsection{Navigation Agent}
For each request $p_k$ with attribute tuple $\mathbf{A}_k$, $\mathcal{N}$ retrieves a stored benign CT and produces a bounding box. TotalSegmentator~\cite{wasserthal2023totalsegmentator} gives a coarse organ mask, and an organ-specific nnU-Net~\cite{isensee2021nnunet} refines it to $m_{\text{ana}}$. A dictionary maps $a_{\text{loc}}$, $a_{\text{inv}}$, and $a_{\text{size}}$ to box sizes measured on real lesions and supplies that context to a pretrained LLM. The LLM samples $b_i=(c_i,w_i,h_i,d_i)$ from this context, $m_{\text{ana}}$, and $\mathbf{A}_k$, turning the text size and invasion attributes into a spatial prior. The box is then clipped to the organ and checked for geometric validity. The synthesis context is $N=\{(x_i,m_{\text{ana},i},b_i)\}$, where $x_i$ is the healthy scan center-cropped to the target organ.

\subsection{Synthesis Agent}
\begin{figure*}[t]
\centering
\includegraphics[width=\textwidth]{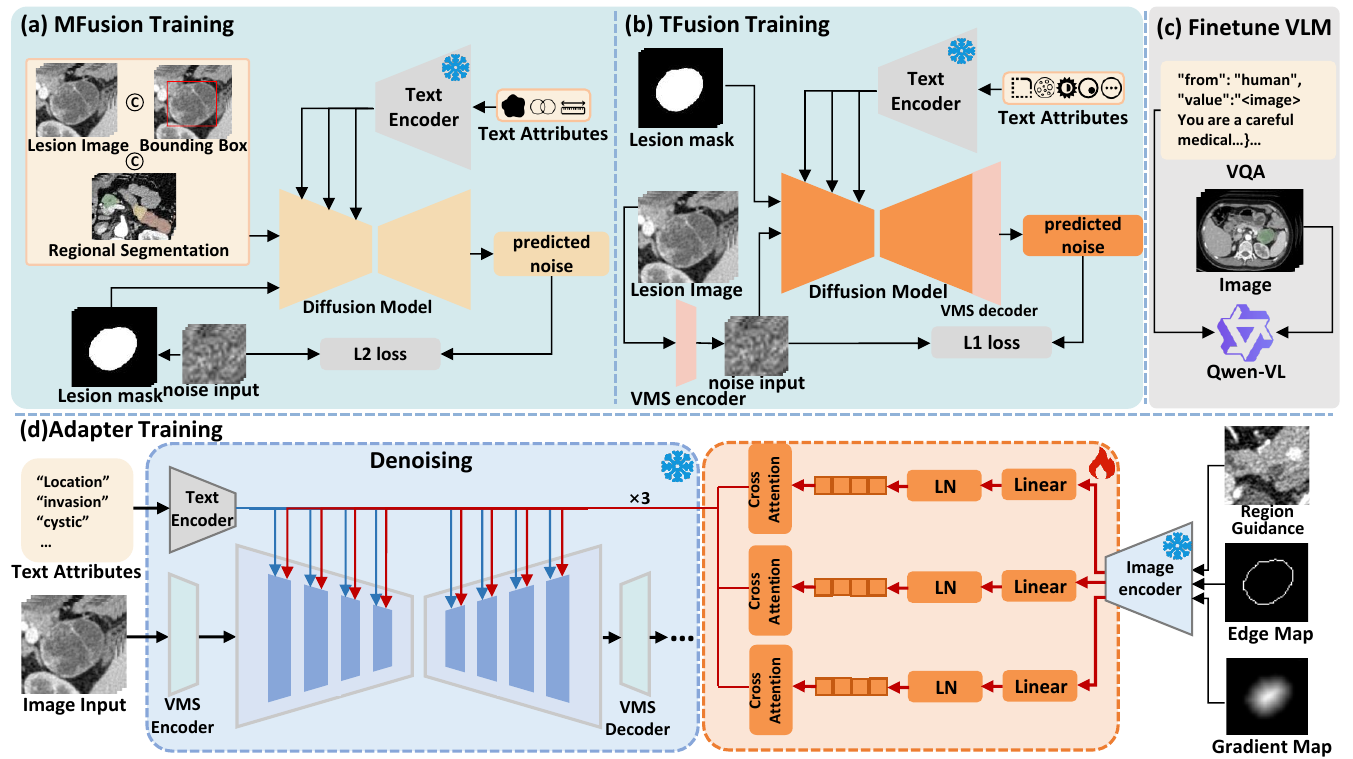}
\caption{Agent preparation of MATS. MFusion and TFusion are trained ahead of the adapters, and the fine-tuned VLM for attribute prediction can be prepared separately.}
\label{fig:synthesis}
\end{figure*}

$\mathcal{S}$ receives the synthesis context together with $\mathbf{A}_k$. As in LesionDiffusion~\cite{lesiondiffusion2025}, mask synthesis (MFusion) is separated from texture rendering (TFusion), so each stage specializes and shape is not entangled with texture (Fig.~\ref{fig:synthesis}). MFusion and TFusion are trained before any adapter. The attribute VLM is fine-tuned independently of both diffusion models.

\noindent\textbf{MFusion.} MFusion is a 3D U-Net diffusion model~\cite{ho2020ddpm}. It is trained on the 425 PNETS training cases. The diffusion target is the expert tumor mask: noise is added to the mask, and the network predicts that noise. The condition is the frozen BiomedCLIP~\cite{biomedclip} embedding $\mathbf{e}_g=E(\mathbf{A}_{\text{geo}})$, the organ-cropped CT, the anatomy mask, and the bounding box. The loss is a squared noise error. Voxels inside the box have weight $\lambda_{\text{in}}{=}5$ and voxels outside have weight $1$. Training uses the 1000-step DDPM schedule; inference uses 200-step DDIM~\cite{song2020ddim}. At inference the same network samples a mask on a healthy scan. The probability volume is binarized at $0.7$, only the largest connected component is kept, and voxels outside $m_{\text{ana}}$ are removed.

\noindent\textbf{TFusion.} TFusion inpaints tumor texture~\cite{lugmayr2022repaint}. A VQ-GAN~\cite{vqvae2017,esser2021vqgan} is first finetuned on CT and then frozen. TFusion diffuses in that latent space. Training uses the 425 real tumor scans: the target is the latent of the tumor CT, and the condition is the expert mask, the host CT, and $\mathbf{e}_t=E(\mathbf{A}_{\text{app}})$. The loss is an absolute noise error, with weight $\lambda_{\text{in}}{=}2$ inside the box and $1$ outside. The decoded patch is blended into the host scan. The schedule is again 1000-step DDPM training and 200-step DDIM inference. At inference the mask is the one accepted from MFusion, and the host is the healthy scan.

\noindent\textbf{Adapters.} Three adapters share one residual cross-attention branch~\cite{ipadapter2023}. With query $\mathbf{Q}$ from the U-Net and text keys and values from $\mathbf{e}_g$ or $\mathbf{e}_t$,
\begin{equation}
\mathbf{O}_{i}=\mathrm{Attn}(\mathbf{Q},\mathbf{K}_{\text{text}},\mathbf{V}_{\text{text}})+\gamma_i\,\mathrm{Attn}(\mathbf{Q},\mathbf{K}_{i},\mathbf{V}_{i}),
\label{eq:adapter}
\end{equation}
where $i\in\{\text{oa},\text{sa},\text{ga}\}$ and $(\mathbf{K}_i,\mathbf{V}_i)$ come from a control feature. The text branch is unchanged. The organ-appearance adapter reads frozen BiomedCLIP features of target-organ parenchyma crops and is applied to TFusion; it has 4.7M trainable parameters and guidance $\gamma_{\text{oa}}{=}\lambda$. The shape adapter reads a retrieved mask exemplar and is applied to MFusion ($\gamma_{\text{sa}}{=}\alpha_s$, about 1.2M parameters). The gradient adapter reads a retrieved Sobel gradient map and is applied to TFusion ($\gamma_{\text{ga}}{=}\alpha_g$, about 1.2M parameters). At layer $l$ the active adapter replaces vanilla cross-attention, $\mathbf{h}^{(l)}=\mathrm{SelfAttn}(\mathbf{h}^{(l-1)})+\mathbf{O}^{(l)}$. Each adapter is trained on the same noise objective as its host model after that host is frozen, using AdamW~\cite{loshchilov2019adamw} on V100-32GB GPUs. MFusion, TFusion, and the VLM stay the PNETS models; only the organ-appearance adapter is fit on a target organ.

\noindent\textbf{Attribute VLM.} Qwen-VL is fine-tuned as a visual question-answering verifier on 2{,}400 labeled attribute checks. On held-out PNET cases its accuracy rises from 41.2\% to 68.7\%. Profiling uses it when a report is missing. Review uses the same checkpoint as a soft score.

\subsection{Review Agent}\label{sec:ra}
Review inserts three checkpoints. Let $s_{\text{vlm}}\in[0,1]$ be the VLM score. A sample may be revised at most $K_{\text{fb}}{=}3$ times; if it still fails, it is discarded and Profiling draws a new request. The VLM score is combined with the deterministic checks below. We do not measure how often a wrong VLM decision accepts a bad sample or rejects a good one.

\noindent\textbf{Feedback 1 (spatial).} This check does not use the IoU between the tumor and the organ. After the MFusion post-process above, every retained voxel already lies inside $m_{\text{ana}}$, including tumors that occupy only a small fraction of the organ. The checkpoint passes when the mask is nonempty and $s_{\text{vlm}}\!\ge\!\tau_1$ for $a_{\text{loc}}$, $a_{\text{inv}}$, and $a_{\text{size}}$. On failure the VLM returns a position offset and a size deviation, and Navigation updates $b_i$ without rerunning segmentation.

\noindent\textbf{Feedback 2 (shape).} The check passes when $s_{\text{vlm}}\!\ge\!\tau_2$ for $a_{\text{shape}}$. On failure the shape adapter retrieves an exemplar from the Shape Bank and reruns MFusion with $\alpha_s\!\propto\!(1-s_{\text{vlm}})$.

\noindent\textbf{Feedback 3 (appearance).} The check passes when $s_{\text{vlm}}\!\ge\!\tau_3$ for $a_{\text{surf}}$, $a_{\text{dens}}$, $a_{\text{vas}}$, and $a_{\text{cyst}}$, and the boundary-gradient $z$-score against real tumors lies in $[-2,2]$. On failure the gradient adapter retrieves a Gradient Bank entry and reruns TFusion with $\alpha_g\!\propto\!(1-s_{\text{vlm}})$. We set $\tau_1{=}0.7$ and $\tau_2{=}\tau_3{=}0.6$. These $\tau$ values are VLM thresholds. The $0.7$ used to binarize MFusion is a separate probability cutoff.

\noindent\textbf{Banks.} The Shape Bank has 256 entries: $k$-means clusters of training masks in the BiomedCLIP feature space, indexed by the dominant attribute tuple. The Gradient Bank has 256 Sobel gradient maps of real tumor texture, indexed by appearance attributes. Retrieval is cosine similarity between the query attributes and the index, with an attribute filter.

\section{Experiments}

\subsection{Experimental Setup}
\noindent\textbf{Data.} In-domain segmentation uses 148 cases drawn from the 174 validation and 141 test cases, disjoint from the 425 synthesis-training cases, and balanced on cystic component, vascularity, and density. Every in-domain number, including ablations, uses this set. Healthy backgrounds are 97 internal scans and 80 NIH scans~\cite{nihpancreas}. Cross-organ evaluation uses five-fold splits of LiTS19~\cite{lits2023} (131 liver tumors), KiTS23~\cite{kits2023} (489 kidney tumors), and MSD-Pancreas~\cite{msd2022} (281 pancreatic tumors).

\noindent\textbf{Baselines.} Copy-Paste pastes real tumor texture onto healthy scans. Rule-based synthesis perturbs HU to imitate vascularity and cystic change. DiffTumor~\cite{difftumor2024} uses its officially released synthetic data. LesionDiffusion~\cite{lesiondiffusion2025} is retrained on PNETS with the same 1:1 budget as MATS. Copy-Paste, Rule-based, and LesionDiffusion follow that shared protocol.

\noindent\textbf{Evaluation.} We train nnU-Net v2~\cite{isensee2021nnunet} and Swin-UNETR~\cite{hatamizadeh2022swinunetr} on the real training set plus an equal number of synthetic cases, on eight V100 GPUs. We report DSC, sample accuracy at IoU $0.5/0.7/0.8$, and attribute-stratified DSC. FID~\cite{heusel2017fid} uses an ImageNet Inception-v3 on tumor-centered axial slices, with the same case count for every method. Cross-organ DSC is the five-fold mean. The reader study uses a two-sided binomial test.

\begin{table*}[t]
\centering
\caption{Segmentation on the 148-case held-out PNETS set. Best in \textbf{bold}; second-best \underline{underlined}. The lower panel is per-attribute DSC (\%) with nnU-Net v2.}
\label{tab:pnet}
\renewcommand{\arraystretch}{1.12}
\setlength{\tabcolsep}{4pt}
\footnotesize
\begin{tabular}{@{}l cccc cccc@{}}
\toprule
\multirow{2}{*}{Method} & \multicolumn{4}{c}{nnU-Net v2} & \multicolumn{4}{c}{Swin-UNETR} \\
\cmidrule(lr){2-5} \cmidrule(lr){6-9}
& DSC$\uparrow$ & Acc@.5$\uparrow$ & Acc@.7$\uparrow$ & Acc@.8$\uparrow$ & DSC$\uparrow$ & Acc@.5$\uparrow$ & Acc@.7$\uparrow$ & Acc@.8$\uparrow$ \\
\midrule
\multicolumn{9}{l}{\textit{Existing methods}} \\
Real only (Baseline) & 58.4 & 68.9 & \underline{56.8} & 39.2 & 50.6 & 56.1 & 41.2 & 21.0 \\
\quad + Copy-Paste & 44.5 & 50.7 & 40.5 & 32.4 & 47.5 & 56.1 & 33.1 & 16.2 \\
\quad + Rule-based & 40.7 & 50.0 & 39.9 & 25.0 & 45.7 & 50.7 & 33.1 & 16.9 \\
\quad + DiffTumor~\cite{difftumor2024} & 54.1 & 62.8 & 49.3 & 33.8 & 49.0 & 59.5 & 36.5 & 17.6 \\
\quad + LesionDiffusion~\cite{lesiondiffusion2025} & 54.6 & 64.9 & 51.4 & 37.2 & 51.8 & 62.2 & 40.5 & 16.2 \\
\midrule
\multicolumn{9}{l}{\textit{Ablation study}} \\
\quad + MATS w/o Profiling & 58.0 & 66.9 & 53.4 & \underline{39.9} & 51.9 & 63.5 & 36.5 & \underline{22.3} \\
\quad + MATS w/o Adapter & 59.0 & \underline{73.0} & 52.7 & 37.8 & 53.7 & \underline{66.9} & \underline{44.6} & 18.9 \\
\quad + MATS w/o Reviewing & \underline{60.9} & 72.3 & 54.7 & 35.1 & \underline{55.3} & 64.9 & 41.9 & 21.0 \\
\midrule
\quad + \textbf{MATS (Full)} & \textbf{62.9} & \textbf{76.4} & \textbf{63.5} & \textbf{43.9} & \textbf{57.0} & \textbf{71.0} & \textbf{45.3} & \textbf{23.7} \\
\bottomrule
\end{tabular}

\vspace{4pt}

\begin{tabular}{@{}l cccc ccc cc@{}}
\toprule
\multirow{2}{*}{Method (nnU-Net v2)} & \multicolumn{4}{c}{Cystic Component} & \multicolumn{3}{c}{Tumor Vascularity} & \multicolumn{2}{c}{Density Variation} \\
\cmidrule(lr){2-5} \cmidrule(lr){6-8} \cmidrule(lr){9-10}
& Solid & Solid-Cyst. & Pred.-Cyst. & Pure Cyst. & Hyper. & Hypo. & Iso. & Homo. & Hetero. \\
\midrule
Baseline & \underline{52.6} & 67.7 & \underline{76.2} & 68.0 & \underline{60.0} & 66.7 & \textbf{38.2} & \underline{58.1} & \underline{58.7} \\
DiffTumor~\cite{difftumor2024} & 46.0 & \underline{75.7} & 73.8 & 65.0 & 54.7 & \underline{71.2} & 32.3 & 56.2 & 52.2 \\
LesionDiffusion~\cite{lesiondiffusion2025} & 46.0 & 73.5 & 74.4 & \underline{71.7} & 56.3 & 64.2 & 31.8 & 53.7 & 55.5 \\
\textbf{MATS (Full)} & \textbf{55.8} & \textbf{78.6} & \textbf{78.6} & \textbf{77.7} & \textbf{63.9} & \textbf{81.6} & \underline{36.7} & \textbf{63.7} & \textbf{62.3} \\
\bottomrule
\end{tabular}
\end{table*}

\begin{figure}[t]
\centering
\includegraphics[width=\columnwidth]{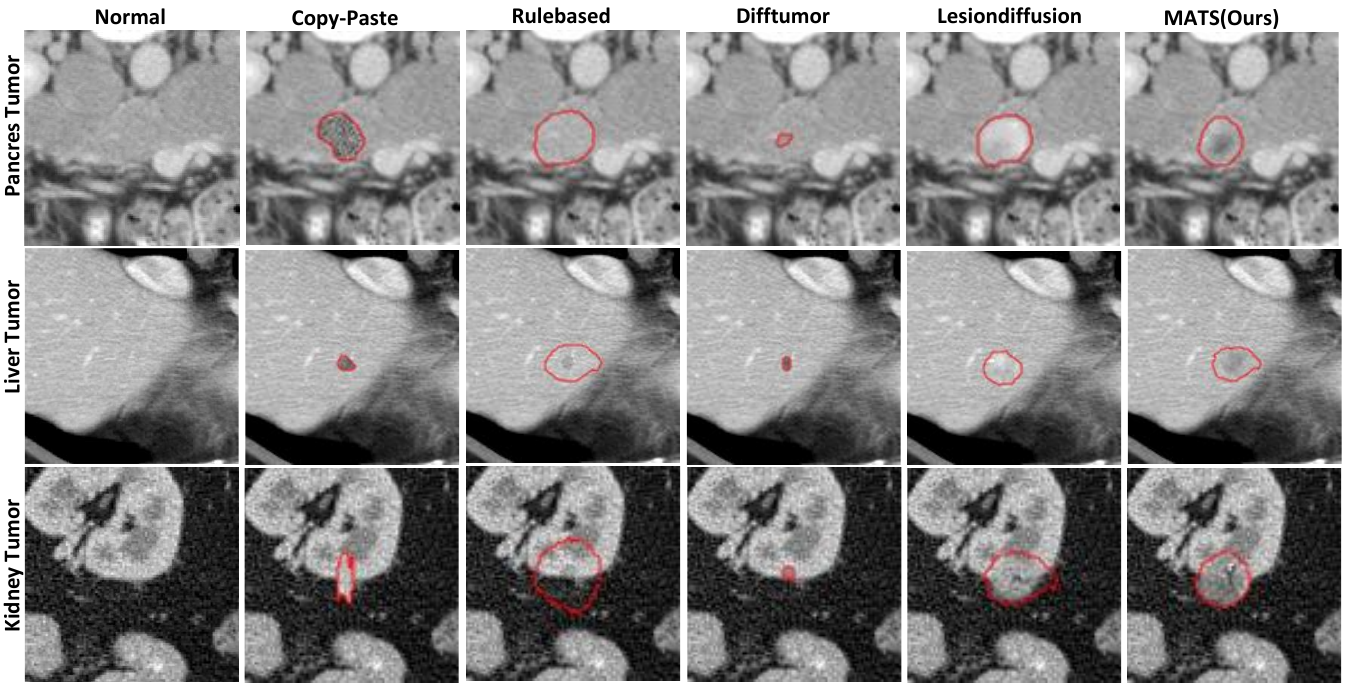}
\caption{Synthetic tumors on pancreas, liver, and kidney. Columns compare methods on the same healthy scan.}
\label{fig:qualitative}
\end{figure}

\begin{figure}[t]
\centering
\includegraphics[width=\columnwidth]{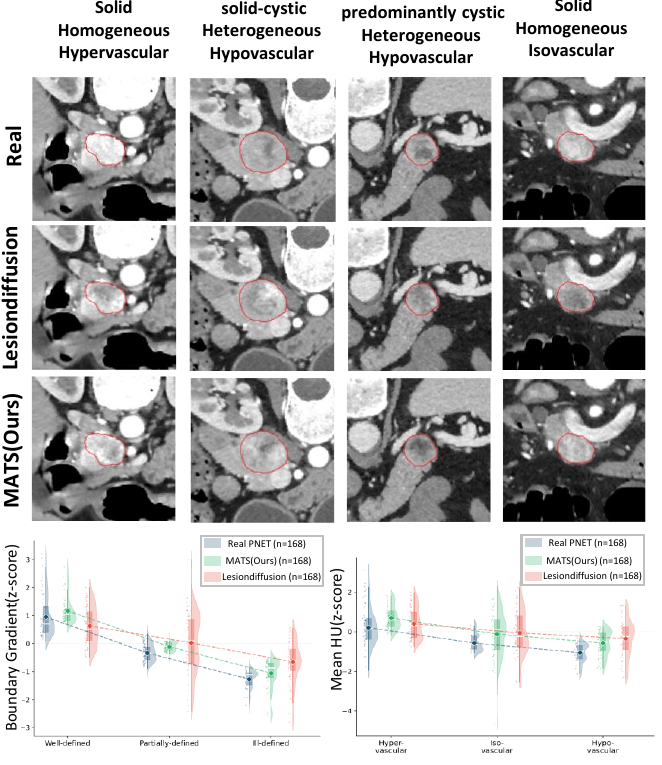}
\caption{Attribute-conditioned examples (left) and $z$-scores of boundary gradient and mean HU against real tumors (right). The curves are those printed in the figure ($n{=}168$ per method) and are not the 148-case segmentation test set.}
\label{fig:controllability}
\end{figure}

\subsection{PNET Synthesis and Segmentation}\label{sec:pnetexp}
Table~\ref{tab:pnet} summarizes segmentation on the 148-case held-out test set. MATS achieves the highest DSC on both backbones (62.9 / 57.0). Notably, most prior generative methods fall below the real-only baseline, suggesting that without distribution-aware control, synthetic augmentation shifts rather than enriches the training distribution; naive copy and rule-based augmentations in particular introduce boundary and intensity artifacts. In contrast, MATS yields consistent gains (+4.5 DSC on nnU-Net v2) through multi-agent orchestration with targeted long-tail attribute planning. The lower panel stratifies DSC by three clinically relevant dimensions: MATS ranks first in 8 of 9 subtypes, verifying the role of $\mathcal{P}$ in overall planning. Our ablations remove one component at a time: \emph{w/o Profiling} replaces profiling with uniform sampling, \emph{w/o Adapter} disables the corrective Shape and Gradient adapters, and \emph{w/o Reviewing} removes the closed-loop review. Removing profiling causes the largest drop in average DSC (62.9$\rightarrow$58.0), confirming the value of long-tail planning, whereas removing the review loop causes the largest drop in the strictest accuracy (Acc@.8: 43.9$\rightarrow$35.1), indicating that review is most valuable for high-fidelity, tightly-aligned tumors; disabling the corrective adapters falls in between (62.9$\rightarrow$59.0).

\subsection{Cross-organ Tumor Synthesis and Segmentation}
All agents are prepared on PNETS and applied to the other organs. Only the $\sim$4.7M-parameter organ-appearance adapter is trained per target organ. Fig.~\ref{fig:qualitative} compares texture on real healthy scans: MATS follows the surrounding intensity, Copy-Paste and Rule-based leave boundary artifacts, DiffTumor is plausible but homogeneous, and LesionDiffusion shows intensity discontinuities. Fig.~\ref{fig:controllability} varies cystic component, vascularity, and density. In those examples the tumor remains inside the organ and the visible texture follows the requested cystic, vascular, and density setting. On boundary-gradient and mean-HU $z$-scores, MATS lies closer to the real-tumor distribution than LesionDiffusion. These two statistics do not measure location, shape, invasion, or size.

Table~\ref{tab:cross} reports cross-organ DSC. MATS is above real-only in all six organ--backbone cells. The largest nnU-Net gain is pancreas ($+2.9$); liver gains are $+0.8$ and $+0.5$. LesionDiffusion is below real-only in all six cells under this single-source protocol. On Swin-UNETR pancreas, DiffTumor reaches 62.2 and MATS reaches 53.7, because the officially released DiffTumor data already contain pancreatic tumors and are in-domain for that cell. The organ-appearance adapter is what carries the PNETS-trained generators onto the target organ.

\begin{table}[t]
\centering
\caption{Cross-organ DSC (\%) under five-fold evaluation. $^\dagger$: officially released synthetic data.}
\label{tab:cross}
\footnotesize
\renewcommand{\arraystretch}{1.12}
\setlength{\tabcolsep}{2.6pt}
\begin{tabular}{l l ccc}
\toprule
Seg.\ Backbone & Synthesis Method & Liver & Kidneys & Pancreas \\
\midrule
\multirow{4}{*}{nnU-Net v2}
& Real only & \underline{80.7} & \underline{88.3} & 61.4 \\
& + DiffTumor$^\dagger$~\cite{difftumor2024} & 73.6 & 82.5 & \underline{63.6} \\
& + LesionDiffusion~\cite{lesiondiffusion2025} & 79.6 & 87.6 & 59.3 \\
& + \textbf{MATS (Ours)} & \textbf{81.5} & \textbf{90.0} & \textbf{64.3} \\
\midrule
\multirow{4}{*}{Swin-UNETR}
& Real only & \underline{75.7} & \underline{83.9} & 46.3 \\
& + DiffTumor$^\dagger$~\cite{difftumor2024} & 71.4 & 76.4 & \textbf{62.2} \\
& + LesionDiffusion~\cite{lesiondiffusion2025} & 72.0 & 76.2 & 46.0 \\
& + \textbf{MATS (Ours)} & \textbf{76.2} & \textbf{85.7} & \underline{53.7} \\
\bottomrule
\end{tabular}
\end{table}

\begin{table}[t]
\centering
\caption{Visual Turing test. Four radiologists, 40 volumes each (20 real, 20 MATS).}
\label{tab:turing}
\footnotesize
\renewcommand{\arraystretch}{1.12}
\setlength{\tabcolsep}{8pt}
\begin{tabular}{lc}
\toprule
Metric & Value \\
\midrule
Classification accuracy$\downarrow$ & 56.3\% \\
Synthetic identified as real$\uparrow$ & 72.5\% \\
Real identified as synthetic & 15.0\% \\
Inter-reader agreement (Fleiss' $\kappa$) & 0.18 \\
\bottomrule
\end{tabular}
\end{table}

\subsection{Generation Quality and Reader Study}
\noindent\textbf{FID.} Each method synthesizes tumors on the same PNET cases. FID~\cite{heusel2017fid} uses an ImageNet-pretrained Inception-v3 on tumor-centered axial slices, so sample count does not confound the comparison. DiffTumor, LesionDiffusion, and MATS score 23.34, 22.52, and 20.97. The margin is modest, and MATS is the closest to real tumors.

\noindent\textbf{Review verifier.} Fine-tuning raises held-out Qwen-VL attribute accuracy from 41.2\% to 68.7\%. This score is combined with the deterministic checks and the budget $K_{\text{fb}}{=}3$. We do not report false-accept or false-rollback rates.

\noindent\textbf{Visual Turing test.} Because PNET is rare, four board-certified radiologists each classified 40 volumes (20 real, 20 MATS) as real or synthetic (Table~\ref{tab:turing}). Accuracy is 56.3\%, not different from 50\% at two-sided binomial $p{=}0.21$. Most synthetic volumes were called real, and inter-reader agreement is low. Near-chance accuracy, low agreement, and that high rate of synthetic volumes called real indicate that these readers did not reliably separate MATS tumors from real PNET CT.

\section{Conclusion}
We presented MATS, a multi-agent framework that plans tumor synthesis from a temperature-smoothed morphological distribution, places the tumor, generates the mask and texture in two diffusion stages, and reviews placement, shape, and appearance in a closed loop. Prepared on PNETS, it improves over real-only segmentation in-domain and on liver, kidney, and pancreas for both backbones, and it is best in 8 of 9 attribute groups. The isovascular group does not improve. The ablations indicate that the planning stage and the review stage contribute beyond simply adding more synthetic volumes. Future work includes a subtype-aware partition of the Gradient Bank for the isovascular presentation, extending agent preparation to partially annotated datasets, and sending downstream segmentation errors back into the review loop.

\bibliographystyle{IEEEtran}
\bibliography{main}

\end{document}